\documentclass{article}
\usepackage{ijcai22}

\usepackage{times}
\usepackage{soul}
\usepackage{url}
\usepackage[hidelinks]{hyperref}
\usepackage[utf8]{inputenc}
\usepackage[small]{caption}
\usepackage{graphicx}
\usepackage{amsmath}
\usepackage{amsthm}
\usepackage{booktabs}
\usepackage[ruled,linesnumbered,vlined]{algorithm2e}

\newcommand{\RNum}[1]{\lowercase\expandafter{\romannumeral #1\relax}}

\usepackage{amsmath}
\usepackage{subcaption}
\usepackage{amssymb}
\usepackage{comment}
\usepackage{bbm}
\usepackage{complexity}
\usepackage{color}

\newtheorem{example}{Example}

\newtheorem{definition}{Definition}

\newcommand{\Real}{\mathbb{R}}
\newcommand{\1}{\mathbbm{1}}

\mathchardef\mhyphen="2D

\title{Branch \& Learn for Recursively and Iteratively Solvable Problems in Predict+Optimize}
\author{
}

\author{
Xinyi Hu$^1$
\and
Jasper C.H. Lee$^2$\and
Jimmy H.M. Lee$^1$\And
Allen Z. Zhong$^1$
\affiliations
$^1$Department of Computer Science and Engineering\\
The Chinese University of Hong Kong, Shatin, N.T., Hong Kong\\
$^2$Department of Computer Science\\
The University of Wisconsin-Madison, WI, USA
\emails
\{xyhu,jlee,zwzhong\}@cse.cuhk.edu.hk,
jasper.lee@wisc.edu
}

\begin{document}

\maketitle

\begin{abstract}
This paper proposes \emph{Branch \& Learn}, a framework for Predict+Optimize to tackle optimization problems containing parameters that are unknown at the time of solving.
Given an optimization problem solvable by a recursive algorithm satisfying simple conditions, we show how a corresponding learning algorithm can be constructed directly and methodically from the recursive algorithm.
Our framework applies also to iterative algorithms by viewing them as a degenerate form of recursion.
Extensive experimentation shows better performance for our proposal over classical and state of the art approaches.
\end{abstract}

\section{Introduction}
In the intersection of machine learning and constrained optimization, the Predict+Optimize framework tackles optimization problems with parameters that are unknown at solving time.
Such uncertainty is common in daily life and industry.
For example, 
retailers need to pick items to restock for maximizing profit, yet consumer demand is a-priori unknown.

The task is to i) predict the unknown parameters, then ii) solve the optimization problem using the predicted parameters, such that the resulting solutions are good even under true parameters.
Traditionally, the parameter prediction uses standard machine learning techniques, with error measures independent of the optimization problem.
Thus, the predicted parameters may in fact lead to a low-quality solution for the (true) optimization problem despite being ``high-quality" for the error metric.  
The Predict+Optimize framework uses the more effective \emph{regret function}~\cite{demirovic2019investigation,elmachtoub2020smart,ali2022divide} as the error metric, capturing the difference in objective (computed under the true parameters) between the estimated and true optimal solutions.
However, the regret function is usually not (sub-) differentiable, and gradient-based methods do not apply.

Prior works have focused on the regime where the optimization problems contain unknown objective and known constraints, and proposed ways to overcome the non-differentiability of the regret.
They can be roughly divided into two approaches:\ \emph{approximation} and \emph{exact}.
The former tries to compute the (approximate) gradients of the regret function or approximations of it.
Elmachtoub {\em et al.\/}~\shortcite{elmachtoub2020smart} propose a differentiable surrogate function for the regret function, while Wilder {\em et al.\/}~\shortcite{wilder2019melding} relax the integral objective in constrained optimization and solve a regularized quadratic programming problem.  
Mandy and Guns \shortcite{mandi2020interior} focus on mixed integer linear programs and propose an interior point based approach. 
While novel, approximation approaches are not always reliable. 
\emph{Exact} approaches exploit the structure of optimization problems to train models without computing gradients.
Demirovi\'{c} {\em et al.\/}~\shortcite{demirovic2019predict+} investigate problems with the ranking property and propose a large neighborhood search method to learn a linear prediction function.
They~\shortcite{demirovic2020dynamic} further extend the method to enable Predict+Optimize for problems amenable to tabular dynamic programming (DP). 

We propose a novel \emph{exact} method for problems solvable with a recursive algorithm (under some restrictions), significantly generalizing the work of Demirovi\'{c} {\em et al.\/}~\shortcite{demirovic2020dynamic}.
By viewing iteration as a special case of recursion with a single branch, our framework applies also to iterative algorithms.
Recall that DP is a special case of recursion, where the recursion structure has overlapping subproblems, enabling the use of memoization to avoid recomputation.
In particular, tabular DP is implemented as an iterative algorithm, computing the table row by row.
While tabular DP is a widely applicable technique, it is not universal in that many naturally recursively solvable problems are not known to have a DP algorithm.
For example, many modern combinatorial search algorithms are still fundamentally based on exhaustive search.
Our work thus subsumes the method by Demirovi\'{c} {\em et al.\/}~\shortcite{demirovic2020dynamic} and extends the Predict+Optimize framework to a much wider class of optimization problems.

Experiments on 2 benchmarks with artificial and real-life data against 9 other learning approaches confirm the superior solution quality, stability, and scalability of our method.

\section{Background}
\label{sec:background}



Without loss of generality, we define an \emph{optimization problem} $P$ as finding:
\begin{equation*}
    x^* = \underset{x}{\arg\min}\ obj(x) \text{ s.t. } C(x)
\end{equation*}
where $x \in \mathbb{R}^d$ is a vector of decision variables, $obj: \mathbb{R}^d \rightarrow \mathbb{R}$ is a function mapping $x$ to a real \emph{objective value} which is to be minimized, and $C$ is a set of constraints over $x$. 
Thus, $x^*$ is an \emph{optimal solution} and $obj(x^*)$ is the \emph{optimal value}.
In this (and prior) work, we focus on the case where only the objective contains uncertainty.
A \emph{parameterized optimization problem (Para-OP)} $P(\theta)$ thus extends the problem $P$ as follows: 
\begin{equation*}
    x^*(\theta) = \underset{x}{\arg\min}\ obj(x, \theta) \text{ s.t. } C(x)
\end{equation*}
where $\theta \in \mathbb{R}^t $ is a vector of parameters.
The objective now depends also on $\theta$. 
When the parameters are known, a Para-OP is just an optimization problem.


\begin{example}\label{example:proj_funding}
Consider a project funding problem to maximize $\sum_{i=1}^4 p_i \cdot x_i$ subject to the constraint $\sum_{i=1}^4 c_i \cdot x_i \leq 3$, where $p$ and $c = [2, 2, 1, 1]$ are two arrays representing the profits and costs of projects. 
The aim is to maximize the total profit under a budget, which is an instance of 0-1 knapsack. 
However, $p$ is usually unknown at decision time, with only some features related to it given, such as proposal scores and the reputation of each applicant: $[(3,3), (4,2.5), (2,4), (3,1)]$.
\end{example}

In \emph{Predict+Optimize} \cite{demirovic2020dynamic}, the true parameters $\theta \in \Real^t$ for a Para-OP are unknown at solving time, and \emph{estimated parameters} $\hat{\theta}$ are used instead. 
Suppose each parameter is estimated by $m$ features.
The estimation will rely on a machine learning model trained over $n$ observations of a training data set $\{(A^1, \theta^1), \dots, (A^n, \theta^n) \}$, where $A^i \in \Real^{t \times m}$ is a \emph{feature matrix} for $\theta^i$, so as to yield a \emph{prediction function} $f:\mathbb{R}^{t \times m} \rightarrow \mathbb{R}^t$ for parameters $\hat{\theta} = f(A)$.

The quality of the estimated parameters $\hat{\theta}$ is measured by the \emph{regret function}, which is the objective difference between the \emph{true optimal solution} $x^*(\theta)$ and the \emph{estimated solution} $x^{*} (\hat{\theta})$ under the true parameters $\theta$.
Formally, we define the regret function $Regret(\hat{\theta}, \theta): \mathbb{R}^t \times \mathbb{R}^t \rightarrow \mathbb{R}_{\geq 0}$ to be: 
\begin{equation*}
    Regret(\hat{\theta},\theta) = obj (x^*(\hat{\theta}), \theta) - obj (x^*(\theta), \theta)  
\end{equation*}
where $obj (x^*(\hat{\theta}), \theta)$ is the \emph{estimated optimal value} and $obj (x^*(\theta), \theta)$  is the \emph{true optimal value}.
Following the empirical risk minimization principle, Elmachtoub, Liang and McNellis~\shortcite{elmachtoub2020decision} choose the prediction function to be the function $f$ from the set of models $\mathcal{F}$ attaining the smallest average regret over the training data:
\begin{equation}
    f^* = \underset{f \in \mathcal{F}}{\arg\min}\  \frac{1}{n}\sum^{n}_{i=1} 
    Regret(f(A^i), \theta^i)  
    \label{eq:ERM}
\end{equation}
For discrete optimization problems, the regret is not (sub) differentiable.
Hence, traditional machine learning algorithms that rely on (sub-)gradients are not applicable.


Demirovi\'{c} {\em et al.\/}~\shortcite{demirovic2020dynamic} study the class $\mathcal{F}$ of linear prediction functions and propose to represent the solution structure of a Para-OP using (continuous) piecewise linear functions.
A \emph{piecewise linear function} $h$ is a real-valued function defined on a finite set of (closed) intervals $\mathbb{I}(h)$ partitioning $\mathbb{R}$.
Each interval $I \in \mathbb{I}(h)$ is associated with a linear function $h[I]$ of the form $h[I](r) = a_I r + b_I$, and the value of $h(r)$ for a real number $r \in \mathbb{R}$ is given by $h[I](r)$ where $r \in I$.
An algebra can be canonically defined on piecewise linear functions~\cite{von1998normal}.
For piecewise linear functions $h$ and $g$, we define pointwise addition as $(h + g)(r) = h(r) + g(r)$ for all $r \in \mathbb{R}$.
Pointwise subtraction, max/min and scalar products are similarly defined.
All five operations can be computed efficiently by iterating over intervals of the operands~\cite{demirovic2020dynamic}. 

In the rest of the paper, we assume that the prediction function $f$ is a \emph{linear mapping of the form $f(A) = A\alpha$} for some $m$-dimensional vector of \emph{coefficients} $\alpha \in \mathbb{R}^m$.


\begin{algorithm}[t]
  \caption{Coordinate Descent}
  \label{alg:coordinate_descent} 
  \KwIn{A Para-OP $P(\theta)$~and a training data set $\{(A^1, \theta^1), \dots, (A^n, \theta^n)\}$}
  \KwOut{a coefficient vector $\alpha \in \mathbb{R}^m$}
  Initialize $\alpha$ arbitrarily and $k \gets 0$\;
  \While{not converged $\wedge$ resources remain}{
    $k \gets (k \mod m) + 1$\;
    Initialize $L$ to be the zero constant function\;
    \For{$i \in [1, 2, \dots, n]$}{
        $(P^i_\gamma, I_0) \gets \texttt{Construct}(P(\theta), k, A^i)$ \; 
        $E^i(\gamma) \gets \texttt{Convert}(P^i_\gamma, I_0)$\;
        $L^i(\gamma) \gets \texttt{Evaluate}(\mathbb{I}(E^i), \theta^i, I_0)$\;
        $L(\gamma) \gets L(\gamma) + L^i(\gamma)$\;
    }
    $\alpha_k \gets {\arg\min}_{\gamma \in \mathbb{R}} L(\gamma)$\;
  }
  return $\alpha$\;
\end{algorithm}

To solve the empirical risk minimization problem in (\ref{eq:ERM}), Demirovi\'{c} {\em et al.\/}~\shortcite{demirovic2020dynamic} propose to update coefficients $\alpha$ of $f$ iteratively via coordinate descent (Algorithm~\ref{alg:coordinate_descent}). 
The algorithm initializes $\alpha$ arbitrarily, and updates the coefficients in a round-robin fashion.
Each iteration (lines 3-10) contains three functions $\texttt{Construct}$, $\texttt{Convert}$, and $\texttt{Evaluate}$.
$\texttt{Construct}$ constructs the optimization problem as a function of the \emph{free coefficient}, fixing the other coefficients in $\alpha$.
$\texttt{Convert}$ takes the parameterized problem and returns a piecewise constant function, representing the estimated optimal values as a function of the free coefficient.
$\texttt{Evaluate}$ subtracts the true optimal value from the predicted optimal values (as a function of the free coefficient) to obtain the regret function, which is also piecewise constant.

As Algorithm~\ref{alg:coordinate_descent} shows, in each iteration (lines 3-10), an coefficient $\alpha_k$ is updated.
Iterating over index $k \in \{1, \dots, m\}$, we replace $\alpha_k$ in $\alpha$ with a variable $\gamma \in \mathbb{R}$ by constructing $\alpha+(\gamma - \alpha_k)e_k$, where $e_k$ is a unit vector for coordinate $k$.
In lines 5-10, we wish to update $\alpha_k$ as:
\begin{equation*}
\begin{aligned}
    \alpha_k \gets &\ 
    \underset{\gamma \in \mathbb{R}}{\arg\min}\  \sum^{n}_{i=1} Regret( A^i e_k\gamma + A^i(\alpha-\alpha_k e_k), \theta^i)
\end{aligned}
\end{equation*}

Let us describe lines 6-8 in more detail.
For notational convenience, let $a^i = A^i e_k \in \mathbb{R}^m$ and $b^i = A^i (\alpha-\alpha_k e_k) \in \mathbb{R}^m$, which are vectors independent of the free variable $\gamma$.
$\texttt{Construct}$ synthesizes the parameterized problem 
\begin{equation*}
\begin{aligned}
    P^i_\gamma &\ \equiv x^*(a^i\gamma + b^i) \\ 
    &\ = \underset{x}{\arg\min}\ obj(x, a^i\gamma + b^i) \text{ s.t. } C(x, a^i\gamma + b^i)
\end{aligned}
\end{equation*}
Sometimes, the parameterized problem can also have an initial domain $I_0 \neq \mathbb{R}$ for $\gamma$.
For instance, we may restrict the estimated profits to be non-negative in Example~\ref{example:proj_funding}.

$\texttt{Convert}$ takes $P^i_\gamma$ to create a function $E^i$ mapping the variable $\gamma$ to the resulting estimated optimal value:
\begin{equation*}
\begin{aligned}
    E^i(\gamma) = obj(x^*(a^i\gamma + b^i), \theta^i)
\end{aligned}
\end{equation*}
In the context of the previous work by Demirovi\'{c} {\em et al.\/}~\shortcite{demirovic2020dynamic}, $E^i$ is always piecewise constant.

Finally, $\texttt{Evaluate}$ computes the (piecewise constant) regret $L^i$
for each interval $I \in \mathbb{I}(E^i)$, i.e. 
\begin{equation*}
\begin{aligned}
 L^i[I] = E^i(\gamma) - obj (x^*(\theta^i), \theta^i) \text{ for some $\gamma \in I$}
\end{aligned}
\end{equation*}
If $\gamma \notin I_0$, the value of $L^i$ is set to a sufficiently large constant to indicate impracticability of the estimation $\hat{\theta} = a^i\gamma + b^i$.

Lastly, line 9 sums all $L^i(\gamma)$ into a piecewise constant total regret $L(\gamma)$ across all training examples, and line 10 minimizes $L(\gamma)$ by simply iterating over each interval of $L(\gamma)$.

While coordinate descent is a standard technique, the key contribution by Demirovi\'{c} {\em et al.\/}~\shortcite{demirovic2020dynamic} is to show how to compute the \verb~Convert~ function for a tabular DP algorithm.
In this paper, we significantly generalize the method by constructing \verb~Convert~ for recursive and iterative algorithms, a much wider class than just DP algorithms.

\section{Recursively Solvable Problems}



This section describes the general form (Algorithm~\ref{alg:Framework_noPara}) of recursive algorithms considered, which in particular captures also all DP algorithms without memoization applied.
This template uses the following two higher-order functions:


\begin{definition}[Map]
Suppose $D, S$ are two sets and $f: D \rightarrow S$ is a function. $\texttt{Map}(f, [d_1, \dots, d_l])$ returns a list $[f(d_1), \dots, f(d_l)]$ where $d_1, \dots, d_l \in D$ and $f(d_i) \in S$.
\end{definition}

\begin{definition}[Reduce]
Suppose $S$ is a set and $\oplus: S \times S \rightarrow S$ is a commutative and associative operation. $\texttt{Reduce}(\oplus, [s_1, s_2, \dots, s_l])$ returns $s_1 \oplus s_2 \oplus \dots \oplus s_l$. 
\end{definition}

The $\texttt{ReSolve}$ function accepts a recursively solvable Para-OP $P$ along with known parameters $\theta$, and returns the optimal value $C^*$ and optimal decisions $x^*$ for $P(\theta)$.
It has several key components.
i) $\texttt{BaseCase}$ determines if $P(\theta)$ is a base case, and if so, $\texttt{BaseResult}$ returns the result of $P(\theta)$.
ii) $\texttt{Extract}$, using the current problem $P$ and the current parameters $\theta$, computes some information $T$ from which the list of subproblems can be determined.
iii) $\texttt{Branch}$, using only $T$, creates and returns a list $PL$ of subproblems, and further computes their corresponding parameters from both $T$ and the current parameters $\theta$.
iv) Each subproblem in the list $PL$ is solved recursively, via $\texttt{map}$ping $\texttt{Resolve}$ to $PL$.
After the recursive calls, the partial results in $RL$ are aggregated by a binary operation $\oplus$ via $\texttt{Reduce}$. 

We also restrict attention to algorithms satisfying:
i) the only arithmetic operations involving the unknown parameters are $+$, $-$, $\max$, $\min$ and multiplication with known constants,
ii) there are no conditionals within \verb~Branch~,
and iii) \verb~BaseCase~ is independent of the parameters $\theta$.

\begin{algorithm}[t]
  \caption{Generic Recursive Solving}
  \label{alg:Framework_noPara} 
  \SetKwProg{Fn}{Function}{:}{}
  \Fn{$\texttt{ReSolve}(P,\theta)$}{
  \If{$\texttt{BaseCase}(P)$}{
    $(x^*(\theta), C^*(\theta)) \gets \texttt{BaseResult}(P,\theta)$\;
  }
  \Else{
    $T \gets \texttt{Extract}(P,\theta)$\;
    $PL \gets \texttt{Branch}(P,\theta,T)$\;
	$RL \gets \texttt{Map}(\texttt{ReSolve}, PL)$\;
	$(x^*(\theta), C^*(\theta)) \gets \texttt{Reduce}(\oplus, RL)$\;
  }
  return $R$
 }
\end{algorithm}

\begin{algorithm}[t]
  \caption{Recursion for 0-1 Knapsack Problem}
  \label{alg:01-knapsack-ReSolve} 
  \SetKwProg{Fn}{Function}{:}{}
  \Fn{$\texttt{ReSolve\_KS}(p, c, n, W, S)$}{
  \If{$n = 0$ or $W \leq 0$}{
    $R \gets \1(W \ge 0) \cdot \sum_{i \in S} p[i]$\;
  }
  \Else{
    $[P_1, P_2] \gets \texttt{Branch\_KS}(p, c, n, W, S)$\;
	$RL \gets \texttt{Map}(\texttt{Resolve\_KS}, PL)$\;
	$R \gets \texttt{Reduce}(\max, RL)$\;
  }
  return $R$
 }
\end{algorithm}

\begin{example}\label{example:recursion}
$\texttt{ReSolve\_KS}$ in Algorithm~\ref{alg:01-knapsack-ReSolve} is an instantiation of $\texttt{ReSolve}$ for solving the project funding (0-1 knapsack) problem in Example~\ref{example:proj_funding}.
$\texttt{ReSolve\_KS}$ takes input $(p, c, n, W, S)$ where $n$ is the number of remaining projects to consider, $W$ is the remaining funding available, and $S$ is the set of selected projects so far.
Initially, $S = \emptyset$.
\verb~Extract_KS~ is a no-op.
$\texttt{Branch\_KS}$ returns $[P_1, P_2]$ where $P_1 = (p, c, n-1, W, S)$ and $P_2 = (p,c,n-1, W-c[n], S\cup\{n\})$ are two subproblems for whether the $n^{th}$ project is selected or not, and \verb~BaseCase~ checks if the total cost exceeds the budget or if no projects are left to consider.
\end{example}

As mentioned, the Algorithm~\ref{alg:Framework_noPara} template also captures iterative algorithms as recursion with a single branch.

\section{The Branch \& Learn Framework}



The proposed \emph{Branch \& Learn} framework methodically transforms a recursive algorithm (Algorithm~\ref{alg:Framework_noPara}) as described in the last section into a Predict+Optimize learning algorithm.
In particular, we adapt the recursive algorithm into the template of Algorithm~\ref{alg:Framework_Para} (\verb~ReLearn~), for use as the \verb~Convert~ function (line 7)---the intellectual core of the Predict+Optimize coordinate descent approach---in Algorithm~\ref{alg:coordinate_descent}.

\begin{algorithm}[t]
  \caption{Generic Recursive Learning}
  \label{alg:Framework_Para} 
  \SetKwProg{Fn}{Function}{:}{}
  \Fn{$\texttt{ReLearn}(P_\gamma, I_0)$}{
    \If{$\texttt{BaseCase}(P_\gamma, I_0)$}{
        $R[I_0] \gets \texttt{BaseResultL}(P_\gamma)$\;
    }
    \Else{
        $T_{\gamma} \gets \texttt{ExtractL}(P_\gamma, I_0)$\;
        \For{each interval $I \in \mathbb{I}(T_{\gamma})$}{
            $PL \gets \texttt{BranchL}(P_{\gamma}, T_{\gamma}[I])$\;
        	$RL \gets \texttt{Map}(\texttt{ReLearn}, PL)$\;
        	$R[I] \gets \texttt{Reduce}(\boxplus, RL)$\;
        }
    }
  return $R$\;
 }
\end{algorithm}


Recall from the Background section that, in the context of the coordinate descent algorithm (Algorithm~\ref{alg:coordinate_descent}), \verb~Convert~ (\verb~ReLearn~ here) takes as input 
i) the problem $P_\gamma$, constructed from $P$ as well as the current training example $(A,\theta)$, which is the problem $P$ expressed as a function of the free coefficient coordinate $\gamma$,
and ii) the domain $I_0$ of $\gamma$, also constructed from $(A,\theta)$ (for example, to ensure basic properties of parameters such as non-negativity).
From these inputs, \verb~ReLearn~ will compute a piecewise constant function, mapping intervals (that partition $I_0$ overall) to estimates of the optimal objective value using $\gamma$ taking any value within that interval.
To construct such a \verb~ReLearn~ procedure, we can simply adapt from the corresponding \verb~ReSolve~ algorithm, and we explain each component of \verb~ReLearn~ here.

Algorithm~\ref{alg:Framework_Para} has component functions \verb~BaseCase~, \verb~BaseResultL~, \verb~ExtractL~, \verb~BranchL~, \verb~Map~ and \verb~Reduce~.
Since \verb~BaseCase~, \verb~Map~ and \verb~Reduce~ do not (directly) involve any parameters, they are the same as the ones in \verb~ReSolve~.
As for \verb~BaseResultL~, \verb~ExtractL~ and \verb~BranchL~, they can be simply obtained from \verb~ReSolve~ by replacing all numerical operations on the parameters with their generalizations to piecewise linear functions.

With these component functions, the structure of \verb~ReLearn~ can then be based on that of \verb~ReSolve~.
\verb~ReLearn~ first checks whether the current subproblem is a base case, and if so, returns the corresponding estimated optimal objective.
Otherwise, it performs the recursion as follows:
i) \verb~ExtractL~ computes information $T_\gamma$ for deciding which subproblems to branch into, where crucially this information will depend on the parameters and hence also on the free coefficient $\gamma$.
Thus the result of \verb~ExtractL~ is a data structure mapping each disjoint subinterval $I$ of $I_0$ to some information $T_\gamma[I]$.
In Algorithm~\ref{alg:Framework_Para}, we denote the set of subintervals as $\mathbb{I}(T_\gamma)$.
ii) With the result of \verb~ExtractL~, we iterate over each $I \in \mathbb{I}(T_\gamma)$, and perform the corresponding \verb~BranchL~, \verb~Map~ and \verb~Reduce~ to get the estimated objective values for each $\gamma \in I$.
Note that, since we map \verb~ReLearn~ instead of \verb~ReSolve~ to $PL$, the estimated objective values may not be constant for all $\gamma \in I$.
Rather, it is a piecewise constant function on $I$, since each subproblem may further divide the subinterval $I$ into smaller subintervals.
\verb~Reduce~, using the canonical piecewise linear extension $\boxplus$ of $\oplus$, handles the further subdivisions as desired.
iii) Finally, the resulting piecewise constant function $R[I]$ is returned.

\begin{algorithm}[t]
  \caption{Parameterized 0-1 Knapsack Problem}
  \label{alg:01-knapsack-ReLearn} 
  \SetKwProg{Fn}{Function}{:}{}

  \Fn{$\texttt{ReLearn\_KS}(p_\gamma, c, n, W, S, I)$}{
  \If{$n = 0$ or $W \leq 0$}{
    $R[I] \gets \sum_{i \in S} p_\gamma[i]$\;
  }
  \Else{
    $[P_1, P_2] \gets \texttt{BranchL\_KS}(p_\gamma, c, n, W, S)$\;
	$RL \gets \texttt{Map}(\texttt{ReLearn\_KS}, PL)$\;
	$R[I] \gets \texttt{Reduce}(\max, RL)$\;
  }
  return $R[I]$
 }
 
\end{algorithm}

\begin{example}\label{example:relearn}
To solve the parameterized 0-1 knapsack problem, the adapted $\texttt{ReLearn\_KS}$ (Algorithm~\ref{alg:01-knapsack-ReLearn}) takes the parameterized problem $P^i_\gamma = (p_\gamma, c, n, W, S)$ and an initial domain $I_0$ as inputs, where $p_\gamma$ is an array of linear functions denoting the estimates of profit parameters as functions of the free coefficient $\gamma$, and $I_0$ is the initial domain depending on $A^i$ such that $p_\gamma \geq 0$ for all $\gamma \in I_0$. 
The components of $\texttt{ReLearn\_KS}$ are essentially the same as those of $\texttt{ReSolve\_KS}$ (Algorithm~\ref{alg:01-knapsack-ReSolve}), except that the addition in line 11 and the $\max$ operation in the $\texttt{Reduce}$ function are replaced by their canonical extensions. 
\end{example}

We also note that, while it is possible to completely formalize the framework into a syntactic transformation, the correctness (or the required restrictions on \verb~ReSolve~) of the formalization will depend highly on the precise programming language used (e.g.~issues of side effect), and require significant low-level work on the level of the grammar of the language, detracting from the key intuitions and principles behind our framework.
For this reason, we do not embark on such formalization in this paper.
Nonetheless, in the next section, we give several case studies to showcase problems that are amenable to our framework and show how to adapt those recursive/iterative algorithms to $\texttt{ReLearn}$ functions, demonstrating the applicability of the proposed framework.


\section{Case Studies}
\label{CaseStudy}

This section gives several case studies of applying the framework.
We first demonstrate, using the examples of 0-1 knapsack (KS) and shortest path (SPP), how our framework can recover the dynamic programming based method by Demirovi\'{c} {\em et al.\/}~\shortcite{demirovic2020dynamic}.
We then showcase our framework on a more complicated iterative algorithm for solving the capacitated minimum cost flow problem (MCFP).
Lastly, to demonstrate the full recursion generality that our framework can handle, we use it on the recursive tree-search algorithm for the $\NP$-hard problem of minimum cost vertex cover (MCVC).


\paragraph{0-1 Knapsack and Shortest Path} 

Tabular DP can be viewed as an iterative algorithm, computing the subproblem table row by row.
In our framework, the \verb~Branch~ operation in \verb~ReSolve~ generates a single subproblem $P'$, which furthermore does not depend on the unknown parameters (only the parameters to $P'$ depend on the unknown parameters to the current problem).
With this perspective, the framework of Demirovi\'{c} {\em et al.\/}~\shortcite{demirovic2020dynamic} is a special case of our method.
An example is 0-1 knapsack with \texttt{ReSolve\_KS} and \texttt{ReLearn\_KS} shown in Algorithms~\ref{alg:01-knapsack-ReSolve} and \ref{alg:01-knapsack-ReLearn} previously.

Another example is the Bellman-Ford algorithm~\cite{cormen2009algorithm} for SPP in a weighted directed graph with potentially negative weights.

\begin{algorithm}[h]
  \caption{Bellman-Ford Algorithm for SPP}
  \label{alg:bellmanford}
  \SetKwProg{Fn}{Function}{:}{}
  \Fn{$\texttt{ReSolve\_SPP}(G^c, D, s, t, N)$}{
    \If{$N = 0$}{
        $R \gets\ D[t]$\;
    }
    \Else{
        $D' \gets \texttt{Extract\_SPP}(G^c, D)$\;
        $[P'] \gets \texttt{Branch\_SPP}(G^c, D', s, t, N-1)$\;
        $R \gets \texttt{ReSolve\_SPP}(P')$\;
    }
    return $R$\;
 }
 
 \Fn{$\texttt{Extract\_SPP}(G^c, D)$}{
    \For{every vertex $u \ne v$ in $V$}{
        $D'[v] = \min( D[v], D[u] + (G^c)_{uv})$\;
    }
    return $D'$
 }
\end{algorithm}

\begin{algorithm}[h]
  \caption{Bellman-Ford Algorithm for Para-SPP}
  \label{alg:para-bellmanford}
  \SetKwProg{Fn}{Function}{:}{}
 
 \Fn{$\texttt{ReLearn\_SPP}(G^c_\gamma, D_\gamma, s, t, N, I_0)$}{
    \If{$N = 0$}{
        $R[I_0] \gets\ D_\gamma[t]$\;
    }
    \Else{
        $D'_\gamma \gets \texttt{ExtractL\_SPP}(G^c_\gamma, D_\gamma, I_0)$\;
        $[P'] \gets \texttt{BranchL\_SPP}(G^c_\gamma, D'_\gamma, s, t, N-1)$\;
        $R[I] \gets \texttt{ReLearn\_SPP}(P', I)$\;
    }
    return $R$\;
 }
 
 \Fn{$\texttt{ExtractL\_SPP}(G^c_\gamma, D_\gamma, I_0)$}{
    \For{every vertex $u \ne v$ in $V$}{
        $D'_\gamma [v] = \min( D_\gamma [v], D_\gamma [u] + (G^c_\gamma)_{uv})$\;
    }
    return $D'_\gamma$
 }
 \end{algorithm}

We instantiate $\texttt{ReSolve}$ for solving SPP to obtain $\texttt{ReSolve\_SPP}$ as follows and shown the pseudocode of \verb~ReSolve_SPP~ and \verb~ReLearn_SPP~ in Algorithms \ref{alg:bellmanford} and \ref{alg:para-bellmanford} respectively.
Suppose $V$ is the set of vertices of the graph. 
The inputs of~$\texttt{ReSolve\_SPP}$ are $(G^c, D, s, t, N)$, where $G^c \in \mathbb{R}^{|V| \times |V|}$ is a matrix with entry $(G^c)_{uv}$ equal to the cost along the directed edge $uv$, $N$ is a counter, $D$ is an array of shortest $(|V|-N-1)$-hop distances from the source to each vertex, and $s$ and $t$ are the source and terminal.
Initially, $N = |V| - 1$ and all elements in $D$ are initialized to a sufficiently large number (signifying infinity), except that $D[s]$ is set to $0$. 
$\texttt{BaseCase\_SPP}$ tests whether $N = 0$, and $\texttt{BaseResult\_SPP}$ returns $D[t]$.
$\texttt{Extract\_SPP}$ computes an array $D'$ where each element $D'[v]$ is computed by iterating over every vertex $u \neq v$ and computing $D'[v] = \min( D[v], D[u] + (G^c)_{uv})$. 
$\texttt{Branch\_SPP}$ returns a single subproblem $[P']$ where $P' = (G^c, D', s, t, N-1)$, on which \verb~ReSolve_SPP~ is called recursively.


We consider the Predict+Optimize setting where the edge costs are unknown.
In the parameterized problem $P_\gamma$, parameterized by the free coefficient $\gamma$ in the coordinate descent, the edge costs are non-negative parameters represented by linear functions of $\gamma$.
Correspondingly, all operations in \verb~ReLearn_SPP~ involving edge costs are replaced by their piecewise linear counterparts (Section~\ref{sec:background}).
$\texttt{ReLearn\_SPP}$ outputs a piecewise constant function of the cost of the estimated shortest path, which can be used to compute the regret.



\paragraph{Capacitated Minimum Cost Flow}

We now apply the B\&L framework to another iteratively solvable problem: MCFP in a directed graph, where at most one edge exists between any two vertices.
Each graph edge has a non-negative capacity and a non-negative cost per unit of flow. 
Given input $K$, we want to find the least cost to route $K$ units of flow from the source $s$ to the terminal $t$.
We consider the parameterized problem where only the edge flow costs are unknown.

We instantiate $\texttt{ReSolve}$ for solving MCFP using the successive shortest path algorithm~\cite{waissi1994network}, to obtain $\texttt{ReSolve\_MCFP}$.
The pseudocode of \verb~ReSolve_MCFP~ and \verb~ReLearn_MCFP~ are shown in Algorithms \ref{alg:ssp} and \ref{alg:para-ssp} respectively.
Suppose the set of vertices is $V$.
The inputs of $\texttt{ReSolve\_MCFP}$ are $(G^p, G^c, F, s, t)$. 
The two matrices $G^{p} \in \mathbb{R}_{\geq 0}^{|V| \times |V|}$ and $G^{c} \in \mathbb{R}_{\geq 0}^{|V| \times |V|}$ represent the edge capacity and the edge (unit) flow cost of the graph respectively.
For the successive shortest path algorithm, we preprocess the graph by adding, for every edge $uv$ in $G$, a reverse edge $vu$ with $0$ capacity and the negated cost $-(G^c)_{uv}$.
The variable $F \in \mathbb{R}_{\geq 0}^{|V| \times |V|}$ is a matrix representing the flow sent along each edge so far.
\verb~BaseCase_MCFP~ tests whether there is no longer a path from $s$ to $t$ with non-zero capacity in $G^{p}$, and \verb~BaseResult_MCFP~ adds the costs of all flows in $F$, i.e.~$R = \sum_{(u,v)} F_{uv} \cdot (G^c)_{uv}$.
\verb~Extract_MCFP~ computes a path $T$ from $s$ to $t$ with lowest cost (per unit flow) in $G^c$. 
\verb~Branch_MCFP~ updates the capacity graph $G^{p}$ and the flow graph $F$ given a path $T$. 
The capacity of each edge in the path $T$ is decreased and the capacity of the reverse edges in $G^{p}$ increased by the largest flow value allowed on $T$.
The flow is also added to $F$.
\verb~ReSolve_MCFP~ is called recursively on these new $G^p$ and $F$ as well as the original $G^c, s$ and $t$.

In the corresponding \verb~ReLearn_MCFP~, the input is a problem $P_\gamma$ parameterized by the free coefficient $\gamma$, and all edge costs (the unknown parameters) are expressed as linear functions of $\gamma$.
The initial domain $I_0$ for $\gamma$ is restricted so that the edge cost estimates are non-negative for all $\gamma \in I_0$.
\verb~ExtractL_MCFP~ adapts from the Bellman-Ford $\texttt{ReLearn\_SPP}$ in the previous case study, which computes a piecewise data structure $T_\gamma$ mapping intervals (for $\gamma$) to different shortest paths (in addition to just the path lengths).
For each interval $I$ of $T_\gamma$, \verb~Branch_MCFP~ constructs a subproblem $P'_\gamma$ by updating $G^{p}$ using $T_\gamma[I]$.
\verb~ReLearn_MCFP~ is recursively called on $P'_\gamma$, until the base case is reached.

\begin{algorithm}[h]
  \caption{Successive Shortest Path Algorithm for MCFP}
  \label{alg:ssp}
  \SetKwProg{Fn}{Function}{:}{}
  \Fn{$\texttt{ReSolve\_MCFP}(G^{p}, G^{c}, F, s, t)$}{
    \If{no path from $s$ to $t$ in $G^{p}$}{
        $R \gets \sum_{(u,v)} F_{uv} \cdot (G^c)_{uv}$\;
    }
    \Else{
        $T \gets \texttt{Extract\_MCFP}(G^{p}, G^{c}, s, t)$\;
        $[P'] \gets \texttt{Branch\_MCFP}(G^{p}, G^{c}, F, T)$\;
        $R \gets \texttt{ReSolve\_MCFP}(P')$\; 
    }
    return $R$\;
 }
 
 \Fn{$\texttt{Extract\_MCFP}(G^{p}, G^{c}, s, t)$}{
    \For{every vertex $v$ in $V$}{
        \For{every vertex $u \ne v$ in $V$}{
            \If{$(G^p)_{uv} > 0$}{
                $D[v] \gets \min( D[v], D[u] + (G^c)_{uv})$\;
                \If{$D[v] = D[u] + (G^c)_{uv}$}{
                    $T[v] \gets u$;
                }
            }
        }
    }
    return $T$;
 }
 
 \Fn{$\texttt{Branch\_MCFP}(G^{p}, G^{c}, F, T)$}{
    $block\_flow = \infty$\;
    \For{every vertex $v$ on $T[]$}{
        $u \gets T[v]$\;
        $block\_flow \gets \min(block\_flow, F_{uv})$\;
    }
    \For{every vertex $v$ on $T[]$}{
        $u \gets T[v]$\;
        $(G^{p \prime})_{uv} \gets (G^p)_{uv} - block\_flow$\;
        $(G^{p \prime})_{vu} \gets (G^p)_{vu} + block\_flow$\;
        $F'_{uv} \gets F_{uv} + block\_flow$\;
    }
    return $(G^{p \prime}, G^c, F', s, t)$\;
 }
\end{algorithm}

\begin{algorithm}[h]
  \caption{Successive Shortest Path Algorithm for Para-MCFP}
  \label{alg:para-ssp}
  \SetKwProg{Fn}{Function}{:}{}
 \Fn{$\texttt{ReLearn\_MCFP}(G^{p}, G^{c}_\gamma, F_\gamma, s, t, I_0)$}{
    \If{no path from $s$ to $t$ in $G^{p}$}{
        $R[I_0] \gets \sum_{I} \sum_{(u,v)} (F_\gamma)_{uv}[I] \cdot (G^c_\gamma)_{uv}$\;
    }
    \Else{
        $T_\gamma \gets \texttt{ExtractL\_MCFP}(G^{p}, G^{c}_\gamma, s, t, I_0)$\;
        \For{each interval $I \in \mathbb{I}(T_{\gamma})$}{
            $[P'] \gets \texttt{BranchL\_MCFP}(G^{p}, G^{c}_\gamma, F_\gamma[I], T_\gamma[I])$\;
            $R[I] \gets \texttt{ReLearn\_MCFP}(P', I)$\; 
        }
    }
    return $R$\;
 }
 
 \Fn{$\texttt{ExtractL\_MCFP}(G^{p}, G^{c}_\gamma, s, t, I_0)$}{
    \For{every vertex $v$ in $V$}{
        \For{every vertex $u \ne v$ in $V$}{
            \If{$(G^p)_{uv} > 0$}{
                 $D_\gamma [v] \gets \min( D_\gamma [v], D_\gamma [u] + (G^c_\gamma)_{uv})$\;
                 \For{each interval $I \in \mathbb{I}(D'_{\gamma})$}{
                    \If{$D_\gamma[v] = D_\gamma[u] + (G^c_\gamma)_{uv}$}{
                        $T_\gamma[I][v] \gets u$\;
                    }
                 }
            }
        }
    }
    return $T_\gamma$;
 }
 
 \Fn{$\texttt{BranchL\_MCFP}(G^{p}, G^{c}_\gamma, F_\gamma[I], T_\gamma[I])$}{
    $block\_flow \gets \infty$\;
    \For{every vertex $v$ on $T_\gamma[I][]$}{
        $u \gets T_\gamma[I][v]$\;
        $block\_flow \gets \min(block\_flow, (F_\gamma[I])_{uv})$\;
    }
    \For{every vertex $v$ on $T_\gamma[I][]$}{
        $u \gets T_\gamma[I][v]$\;
        $(G^{p \prime})_{uv} \gets (G^p)_{uv} - block\_flow$\;
        $(G^{p \prime})_{vu} \gets (G^p)_{vu} + block\_flow$\;
        $F'_\gamma[I]_{uv} \gets (F_\gamma[I])_{uv} + block\_flow$\;
    }
    return $(G^{p \prime}, G^c_\gamma, F'_\gamma[I], s, t, I)$\;
 }
\end{algorithm}

\paragraph{Minimum Cost Vertex Cover}

Our last example is the minimum cost vertex cover (MCVC) problem, where we show how to apply our framework to a (non-degenerate) recursive algorithm (with multiple branches).
We show the pseudocode of \verb~ReSolve_MCVC~ and \verb~ReLearn_MCVC~ in Algorithms \ref{alg:mcvc} and \ref{alg:para_mcvc} respectively. 
Given a graph $G = (V,E)$, there is an associated \emph{cost} $c \in \Real^{|V|}$ denoting the cost of picking each vertex.
The costs are unknown parameters.
The goal is to pick a subset of vertices, minimizing the total cost, subject to the constraint that all edges need to be covered, namely at least one of the two vertices on an edge needs to be picked.
This problem is relevant in applications such as building public facilities.
Consider, for example, the graph being a road network with edge values being traffic flow, and we wish to build speed cameras at intersections with minimum cost, while covering all the roads.

The recursive algorithm \verb~ReSolve_MCVC~ takes input $(G,c,\ell,n,chosen)$, where $G,c,\ell$ are as before, $n$ is the number of levels of (binary) recursion remaining and $chosen$ is the current list of chosen vertices.
\verb~BaseCase_MCVC~ checks if $n = 0$, and \verb~BaseResult_MCVC~ returns the total cost of vertices in $chosen$ if all edges in $G$ are covered, and returns infinity otherwise.
\verb~Extract_MCVC~ is a no-op.
\verb~Branch_MCVC~ creates two subproblems $(G,c,\ell,n-1,chosen)$ and $(G,c,\ell,n-1,chosen \cup \{n\})$, i.e.~choosing vertex $n$ or not, and \verb~Reduce~ takes the min of the two options.

Correspondingly, \verb~ReLearn_MCVC~ replaces all arithmetic and $\min$ operations by piecewise linear counterparts.


\begin{algorithm}[h]
  \caption{Branching Algorithm for MCVC}
  \label{alg:mcvc}
  \SetKwProg{Fn}{Function}{:}{}
  \Fn{$\texttt{ReSolve\_MCVC}(G, c, \ell, n, chosen)$}{
    \If{$n=0$}{
        \If{the edges in $G$ are all covered}{ 
            $R \gets 0$\;
            \For{every vertex $v$ in $chosen[]$}{
                $R \gets R + c[v]$\;
            }
            return $R$\;
        }
        \Else{
            return $\infty$\;
        }
    }
    \Else{
        $[P_1, P_2] \gets \texttt{Branch\_MCVC}(G, c, \ell, n, chosen)$\;
        $[R_1, R_2] \gets \texttt{Map}(\texttt{ReSolve\_MCVC}, [P_1, P_2])$\;
        $R \gets \texttt{Reduce}(\boldsymbol{min}, [R_1, R_2])$\;
    }

    return $R$\;
 }

 \Fn{$\texttt{Branch\_MCVC}(G, c, \ell, n, chosen)$}{
    $P_1 \gets (G,c,\ell,n-1,chosen)$\;
    $P_2 \gets (G,c,\ell,n-1,chosen \cup \{n\})$\;
    return $(P_1, P_2)$\;
 }
 
\end{algorithm}

\begin{algorithm}[h]
  \caption{Branching Algorithm for Para-MCVC}
  \label{alg:para_mcvc}
  \SetKwProg{Fn}{Function}{:}{}
  \Fn{$\texttt{ReLearn\_MCVC}(G, c_\gamma, \ell_\gamma, n, chosen_\gamma, I_0)$}{

    \If{$n=0$}{
        \If{the edges in $G$ are all covered}{ 
            $R[I_0] \gets 0$\;
            \For{every vertex $v$ in $chosen_\gamma[]$}{
                $R[I_0] \gets R[I_0] + c_\gamma[v]$\;
            }
            return $R[I_0]$\;
        }
        \Else{
            $R[I_0] \gets \infty$\;
            return $R[I_0]$\;
        }
    }
    \Else{
        $[P_1, P_2] \gets \texttt{BranchL\_MCVC}(G, c_\gamma, \ell_\gamma, n, chosen_\gamma, I_0)$\;
        $[R_1, R_2] \gets \texttt{Map}(\texttt{ReSolve\_MCVC}, [P_1, P_2])$\;
        $R \gets \texttt{Reduce}(\boldsymbol{min}, [R_1, R_2])$\;
    }

    return $R$\;
 }

 \Fn{$\texttt{BranchL\_MCVC}(G, c_\gamma, \ell_\gamma, n, chosen_\gamma, I_0)$}{
    $P_1 \gets (G,c_\gamma,\ell_\gamma,n-1,chosen_\gamma, I_0)$\;
    $P_2 \gets (G,c_\gamma,\ell_\gamma,n-1,chosen_\gamma \cup \{n\}, I_0)$\;
    return $(P_1, P_2)$\;
 }
 
\end{algorithm}

\section{Experimental Evaluation}

Our experiments are on MCFP and MCVC---problems that the previous DP-based method cannot handle.
We use both artificial and real-life data on real-life graphs.
For MCFP, we use USANet~\cite{lucerna2009efficiency}, with 24 vertices and 43 edges, and G\'{E}ANT~\cite{WinNT}, with 40 vertices and 61 edges.
For the NP-hard problem of MCVC, we use two smaller graphs from the Survivable Network Design Library~\cite{SNDlib10}: POLSKA, with 12 vertices and 18 edges, and PDH, with 11 vertices and 34 edges.
In MCFP, the edge costs are unknown, and the capacities are sampled from $\lbrack 10, 50 \rbrack$.
We set the flow value to be 20, and select random source and sink.
In MCVC, the costs are unknown.

We run $30$ simulations for each problem configuration.
In each simulation, we build datasets consisting of $n \in \{100, 300\}$ pairs of (feature matrix, parameters).
In the artificial and real-life datasets, each parameter has 4 and 8 features respectively.

The artificial dataset for both of the two problems is generated as follows.
Each feature is a 4-tuple $\Vec{a}_{uv} = (a_{uv1},a_{uv2},a_{uv3},a_{uv4})$, where $a_{uv1} \in \lbrace 1,2,\ldots, 7 \rbrace$ represents the day of the week, $a_{uv2} \in \lbrace 1,2,\ldots, 30 \rbrace$ represents the day of the month, and $a_{uv3},a_{uv4} \in \lbrack 0,360 \rbrack$ represent the meteorology index and road congestion respectively. 
The true parameters are generated by $10*sin(a_{uv1})*sin(a_{uv2})+100*sin(a_{uv3})*sin(a_{uv4})+C$, where $C$ is a large positive constant to ensure the value of each learned parameter is positive. 
We use such nonlinear mapping to compare the performance of our proposed methods and that of other methods. 

Given that we are unable to find datasets specifically for the MCFP and MCVC problems, we follow the experimental approach of Demirovic \emph{et al.}~\cite{demirovic2019investigation,demirovic2019predict+,demirovic2020dynamic} and use real data from a different problem (the ICON scheduling competition) as numerical values required for our experiment instances.

We use a 70\%/30\% training/testing data split.
We use the \textit{scikit-learn} library \cite{sklearn_api} to implement LR, $k$-NN, CART and RF, and \textit{or-tools}~\cite{ortools} as the problem solver in SPOT and SPO Forest. 
As for the parameters tuning, we try different settings for k-nearest neighbors (k-NN), Random forest (RF), SPO tree (SPOT), and SPO Forest. 
In k-NN, we try the regression model with $k \in \lbrace 1,3,5 \rbrace$. 
As for RF, we try different numbers of trees in the forest $n\_estimator \in \lbrace 10, 50, 100 \rbrace$. 
We tune the maximum depths of the tree $max\_depth \in \lbrace 1, 3, 10, 100 \rbrace$ and the minimum weights per node $min\_weights \in \lbrace 5, 20, 30 \rbrace$ for SPOT. 
For SPO Forest, we tune two parameters: the maximum depth of each tree $max\_depth \in \lbrace 1, 3, 10, 100  \rbrace$ and the number of trees in the forest $n\_estimator \in \lbrace 10, 50, 100 \rbrace$.

\paragraph{Solution Quality}

\begin{table*}[h]
\centering
\resizebox{1\textwidth}{!}{
\begin{tabular}{|l||cc|cc||cc|cc|}
\hline
 & \multicolumn{4}{c||}{Artificial Dataset}                                                               & \multicolumn{4}{c|}{Real-life Dataset}                                                  \\ \cline{2-9} 
                      & \multicolumn{2}{c|}{USANet}                         & \multicolumn{2}{c||}{G\'{E}ANT}                      & \multicolumn{2}{c|}{USANet}                    & \multicolumn{2}{c|}{G\'{E}ANT}                  \\ \cline{2-9} 
Size                      & 100                      & 300                      & 100                    & 300                    & 100                    & 300                   & 100                  & 300                  \\ \hline
B\&L                  & \textbf{1687.10±729.36} & \textbf{1699.24±688.96} & \textbf{733.78±294.23} & \textbf{732.99±270.88} & \textbf{141.37±128.52} & \textbf{122.66±85.52} & \textbf{72.43±63.58} & \textbf{68.44±50.76} \\
LR                    & 1795.02±792.63          & 1749.24±691.16          & 765.13±345.07          & 743.20±289.98          & 177.55±154.53          & 141.00±92.60          & 84.41±68.03          & 88.66±63.21          \\
$k$-NN                  & 1783.50±783.84          & 1791.61±715.92          & 801.65±350.87          & 759.87±278.67          & 285.30±206.60          & 223.95±132.56         & 127.73±79.25         & 121.75±48.45         \\
CART                  & 2529.74±891.06          & 2436.37±922.07          & 1310.53±422.75        & 1260.65±347.26        & 339.41±219.52          & 313.36±175.66         & 157.82±62.13         & 194.62±88.12         \\
RF                    & 1783.84±750.26          & 1707.96±670.23          & 766.73±322.94          & 771.19±314.24          & 197.25±162.16          & 138.02±87.96          & 88.25±58.49          & 103.92±56.31         \\
SPO                   & 2193.87±880.07           & 2071.87±739.14           & 1011.65±322.98         & 937.57±252.87          & 204.89±185.62          & 139.55±93.63          & 82.48±71.62          & 84.18±57.95          \\
QPTL                  & 2220.47±850.36           & 2244.40±854.32           & 1063.96±355.06         & 1093.78±282.53         & 259.94±236.54          & 212.83±172.12         & 84.60±71.89          & 102.30±61.28         \\
IntOpt                & 1796.33±756.94           & 1754.77±701.96           & 778.18±303.49          & 762.99±293.06          & 200.69±174.94          & 140.98±88.58          & 82.42±68.32          & 87.61±58.05          \\
SPOT                  & 1743.68±754.79          & 1723.44±695.73          & 1487.71±879.45        & 1499.39±845.48        & 185.68±156.35          & 153.47±106.54         & 143.57±107.88        & 136.16±85.02         \\
SPO Forest            & 1777.86±689.84          & 1736.28±691.08          & 745.43±344.07          & 747.60±292.79          & 178.35±144.90          & 145.52±102.28         & 135.98±101.46        & 132.07±77.23         \\ \hline \hline
Average TOV    & 10825.18±921.41         & 10835.36±1038.05       & 9831.18±3318.39      & 9784.31±3391.45      & 6831.07±1044.27      & 6660.78±872.37       & 6459.94±2049.60      & 6026.91±2049.46      \\ \hline
\end{tabular}}
\caption{Mean regrets and standard deviations for MCFP with unknown parameters}
\label{table:MC_avgReg}
\end{table*}

\begin{table*}[h]
\centering
\resizebox{0.9\textwidth}{!}{
\begin{tabular}{|l||cc|cc||cc|cc|}
\hline
 & \multicolumn{4}{c||}{Artificial Dataset}                                                               & \multicolumn{4}{c|}{Real-life Dataset}                                                  \\ \cline{2-9} 
                      & \multicolumn{2}{c|}{POLSKA}                         & \multicolumn{2}{c||}{PDH}                      & \multicolumn{2}{c|}{POLSKA}                    & \multicolumn{2}{c|}{PDH}                  \\ \cline{2-9} 
Size                      & 100                      & 300                      & 100                    & 300                    & 100                    & 300                   & 100                  & 300                  \\ \hline
B\&L                & \textbf{109.45±15.42} & \textbf{110.04±6.55} & \textbf{50.40±8.60} & \textbf{53.71±5.62} & \textbf{2.22±1.36} & \textbf{2.18±0.49} & \textbf{7.57±6.04} & \textbf{5.51±2.62} \\
LR                & 115.69±15.59          & 116.36±8.60          & 55.15±9.98          & 56.59±6.70          & 3.56±2.07          & 3.09±0.91          & 8.32±5.86          & 6.10±2.67          \\
$k$-NN              & 123.06±17.26          & 116.97±11.26         & 56.58±8.77          & 58.36±6.16          & 5.16±2.13          & 5.02±1.00          & 11.77±7.23         & 9.78±2.91          \\
CART              & 117.23±14.88          & 123.61±13.20         & 89.68±15.95         & 89.40±8.91          & 5.47±2.29          & 5.38±1.21          & 16.92±7.64         & 11.70±3.51         \\
RF                & 116.38±14.75          & 117.13±11.75         & 58.75±9.88          & 56.84±6.62          & 4.17±1.83          & 3.91±0.92          & 13.45±7.70         & 7.76±3.36          \\
SPO               & 117.50±16.94          & 116.72±9.14          & 78.74±14.98         & 69.46±6.63          & 2.94±1.54          & 2.78±0.82          & 10.00±7.97         & 6.58±3.07          \\
QPTL              & 118.14±18.07          & 117.52±9.14          & 81.70±14.83         & 79.31±5.72          & 2.58±1.37          & 2.57±0.62          & 10.89±6.49         & 9.28±3.99          \\
IntOpt            & 119.52±16.04          & 118.44±9.70          & 69.66±14.43         & 66.12±8.44          & 2.55±1.32          & 2.30±0.51          & 10.93±6.56         & 8.79±3.93          \\
SPOT              & 118.90±17.62          & 117.55±9.49          & 56.51±8.95          & 57.70±6.77          & 2.68±1.31          & 2.57±0.65          & 14.12±6.73         & 11.19±4.13         \\
SPO Forest              & 119.00±18.72          & 117.25±10.95         & 55.07±8.98          & 56.85±6.00          & 2.86±1.42          & 2.66±0.64          & 13.11±7.40         & 11.45±3.62         \\ \hline \hline
Average TOV               & 649.00±21.02          & 654.82±11.83         & 817.76±26.29        & 822.72±16.22        & 321.14±16.11       & 317.96±6.96        & 502.17±30.42       & 503.57±12.39 \\ \hline
\end{tabular}}
\caption{Mean regrets and standard deviations for MCVC with unknown parameters}
\label{table:MCVC_avg2S}
\end{table*}

We design graph-specific distributions for taking a random source and a random sink for the minimum cost flow problem (MCFP), with the goal of making sure that the path between the source and the sink is not too short (e.g.~length 1).
In USANet, 
we randomly choose the source from vertices $\{1,2,3,4,5\}$ and the sink from vertices $\{20,21,22,23,24\}$.
In G\'{E}ANT, the source and sink are randomly selected from all the points with zero in-degree and zero out-degree respectively.

Table \ref{table:MC_avgReg} reports the mean regrets and their standard deviations for each method on MCFP.
We observe that B\&L achieves the best performance in all cases. 
On the artificial dataset, all algorithms achieve similar performance on USANet except CART, while both CART and SPOT perform poorly on G\'{E}ANT with the real-life dataset. 
With the real-life dataset, B\&L shows the most significant advantages.
Compared with other methods, B\&L obtains 20.38\%-58.35\% ($n = 100$) and 11.13\%-60.86\% ($n = 300$) smaller regret on USANet, and 12.12\%-54.11\% ($n=100$) and 18.70\%-64.83\% ($n=300$) smaller regret on G\'{E}ANT.

We also report the average True Optimal Values (TOV) to compare the relative error on the artificial and real-life datasets.
We observe that all methods achieve smaller relative error with real-life data than with artificial data.
B\&L, for example, achieves 7.46\%-15.68\% and 1.12\%-2.07\% relative error with artificial and real-life data respectively.
This is consistent with how the artificial dataset is purposefully designed to be highly non-linear, and thus harder to learn. 
Nevertheless, B\&L still achieves the smallest regret.

Table \ref{table:MCVC_avg2S} shows the mean regrets and standard deviations in the MCVC experiment.
We can see that B\&L achieves the best performance in all cases.
B\&L obtains 12.95\%-40.59\% ($n=100$) and 9.05\%-44.74\% ($n=300$) smaller regret in POLSKA, and 5.22\%-40.47\% ($n=100$) and 9.71\%-47.11\% ($n=300$) in PDH. 
B\&L achieves 6.16\%-16.86\% relative error in the artificial dataset, and 0.68\%-1.51\% relative error in the real-life dataset.

\paragraph{Stability}

\begin{figure*}[h]
\centering
\begin{subfigure}{0.45\textwidth}
  \centering
  \includegraphics[width=1\textwidth]{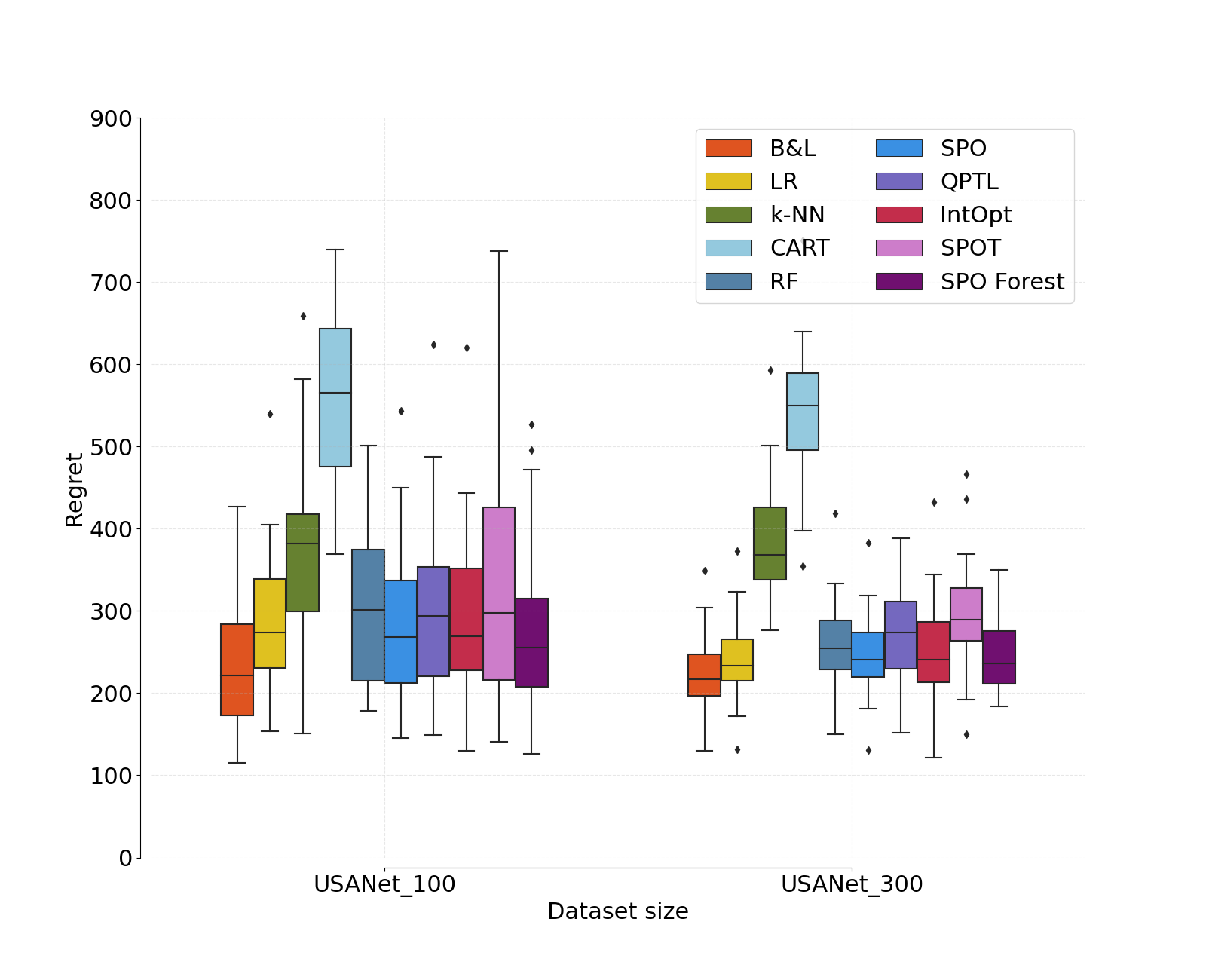}
    \caption{Regret of MCFP on USANet}
    \label{MCFP_USANet_bp}
\end{subfigure}
\begin{subfigure}{0.45\textwidth}
  \centering
  \includegraphics[width=1\textwidth]{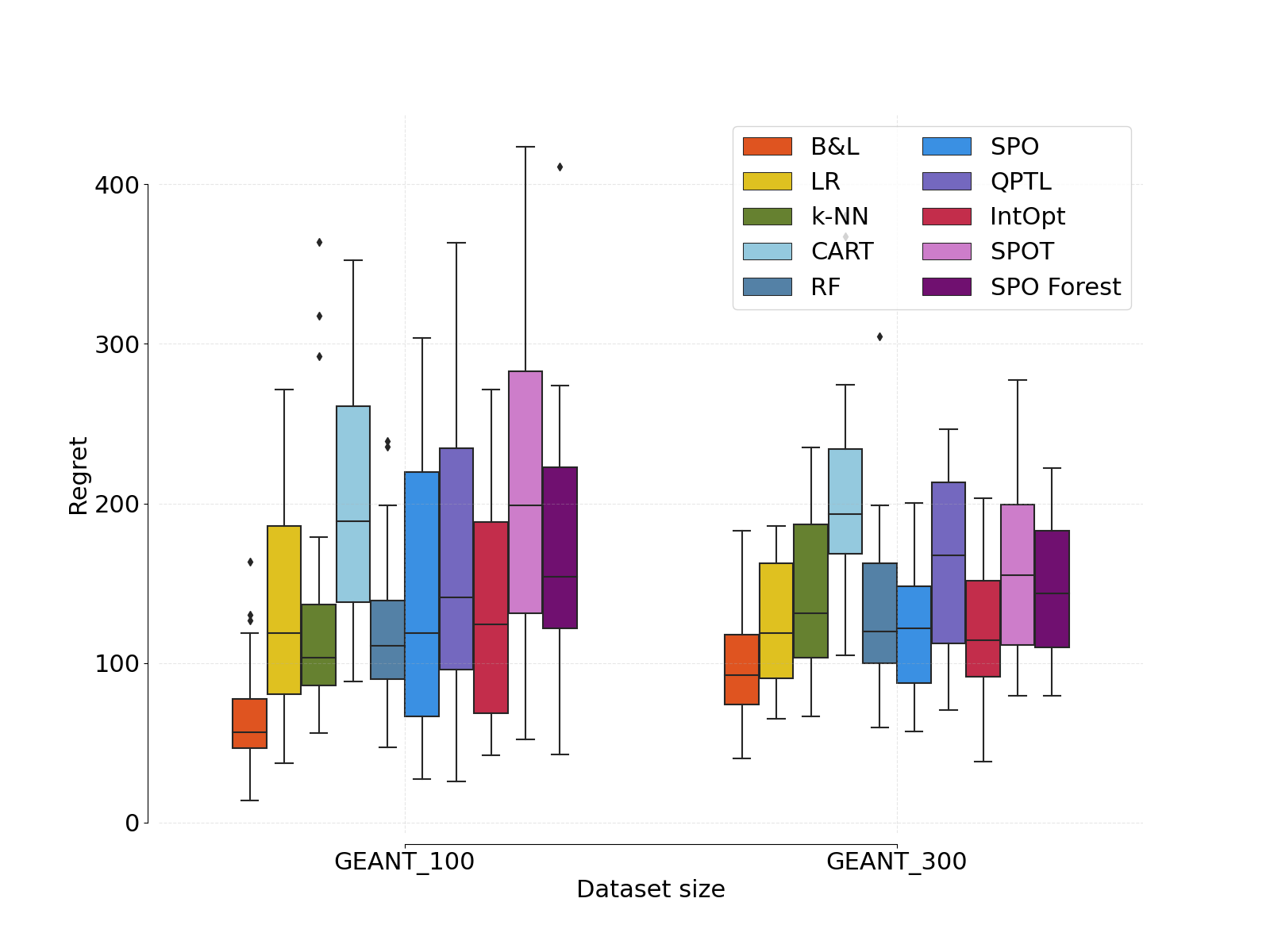}
  \caption{Regret of MCFP on GEANT}
    \label{MCFP_GEANT_bp}
\end{subfigure}
\caption{Boxplots of MCFP on real-life dataset.}
\label{MCFP_bp}
\end{figure*}

\begin{figure*}[h]
\centering
\begin{subfigure}{0.45\textwidth}
  \centering
  \includegraphics[width=1\textwidth]{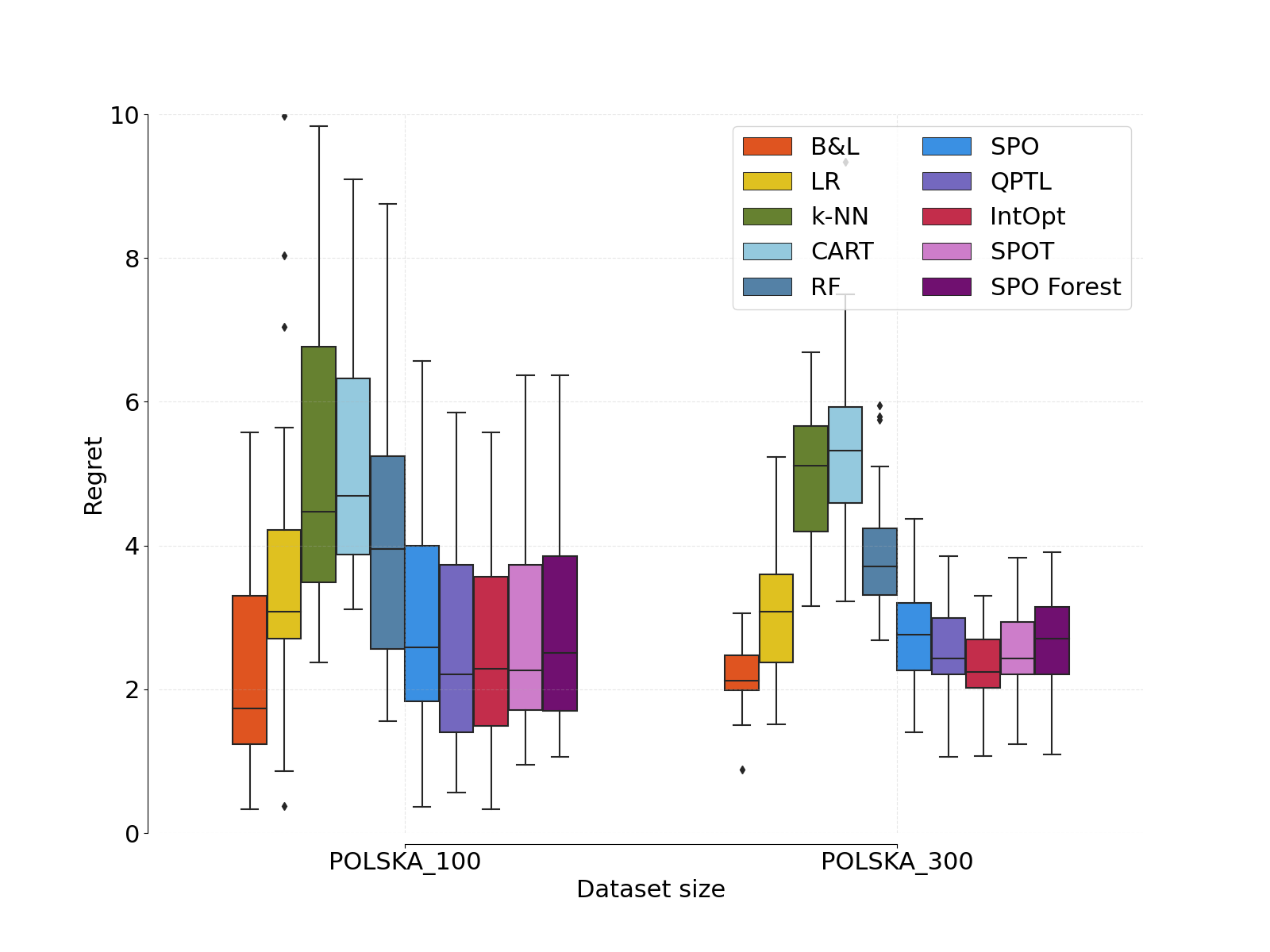}
    \caption{Regret of MCVC on POLSKA}
    \label{MCVC_POLSKA_bp}
\end{subfigure}
\begin{subfigure}{0.45\textwidth}
  \centering
    \includegraphics[width=1\textwidth]{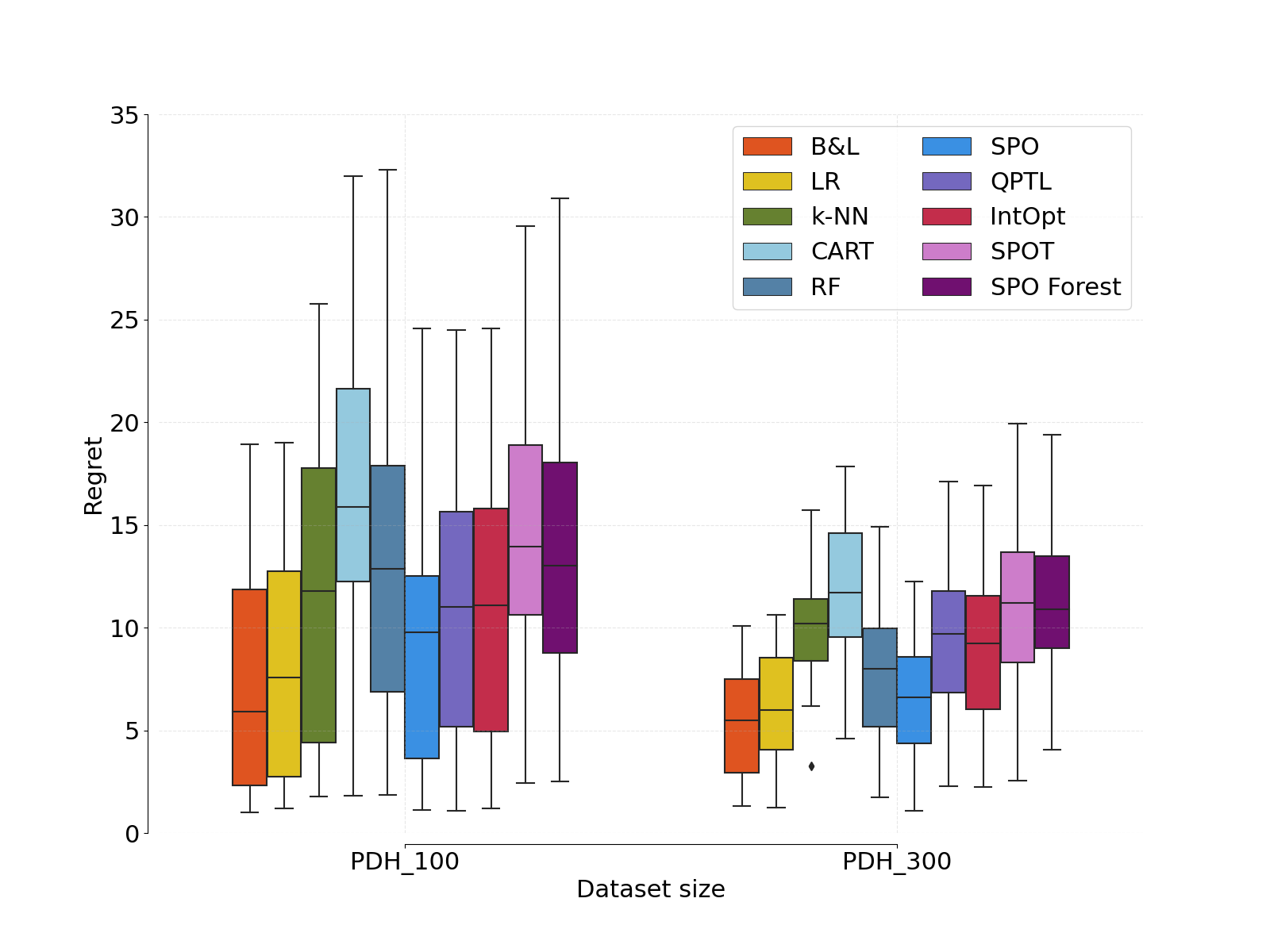}
    \caption{Regret of MCVC on PDH}
    \label{MCVC_PDH_bp}
\end{subfigure}
\caption{Boxplots of MCVC on real-life dataset.}
\label{MCVC_bp}
\end{figure*}

To better understand the distribution of regrets across the different learning models in our experiments, we report the associated box plots in this section.
For MCFP, when the source and sink are randomly selected, the objective value and the regret may have big deviations.
Therefore, for more accurate observations on the performance differences between our proposed methods and other methods, we fix the source as 0 and the sink as 23 in USANet and fix the source as 0 and the sink as 31 in G\'{E}ANT.
Figures \ref{MCFP_bp} and \ref{MCVC_bp} show the box plot of each model's solution quality on MCFP and MCVC respectively.
We vary the total dataset size $n \in \lbrace 100, 300\rbrace$, and note that the box plot for each configuration is across the 30 independent simulations.

Figures \ref{MCFP_USANet_bp} and \ref{MCFP_GEANT_bp} show the box plots of regret of MCFP on USANet and G\'{E}ANT respectively.
We can see that the median regret values, the dispersion, and the outliers of all models tend to decline as the data set size increases. 
Among the two set sizes, we observe that B\&L has the lowest median lines and the shortest interquartile ranges, suggesting that it can achieve the smallest regret as well as good stability.
In addition, linear regression performs second best among other models, since it has the same linear prediction function as the proposed algorithm.


Figures \ref{MCVC_POLSKA_bp} and \ref{MCVC_PDH_bp} show the box plots of regret of MCVC on POLSKA and PDH respectively.
All the methods have large interquartile ranges when the set size is 100, and have shorter interquartile ranges when the set size is 300.
When the set size is 100, although B\&L has large interquartile ranges, the median lines are much lower than any other methods.
When the set size is 300, the median lines of B\&L are lower than other methods, and the interquartile ranges are shorter than or similar to other methods.


\paragraph{Scalability/Runtime}

Learning using regression or approximate methods is fast, but these methods sacrifice the accuracy of the learned model, which is the motivation for the line of work on Predict+Optimize.
On the other hand, many optimization problems are expensive to solve.  Although exact methods achieve lower regret, their runtime scalability can be an issue since their learning process requires solving optimization problems multiple times.
For a fair comparison, therefore, we compare B\&L only with other exact methods: SPOT and SPO Forest.
Table \ref{table:MCFP_runtime} shows the average runtime across 30 simulations for different cases.
Overall, we observe that B\&L scales at least as well as SPOT and SPO Forest.
We note that the runtime of SPOT and SPO Forest becomes quite large in G\'{E}ANT, POLSKA and PDH when the total set size is 300, while B\&L maintains good runtime behavior.

\begin{table}[h]
\centering
\resizebox{0.48\textwidth}{!}{
\begin{tabular}{|l||cc|cc||cc|cc|}
\hline
 & \multicolumn{4}{c||}{Minimum cost flow problem} & \multicolumn{4}{c|}{Minimum cost vertex covering problem}     \\ \cline{2-9} 
 & \multicolumn{2}{c|}{USANet}                         & \multicolumn{2}{c||}{G\'{E}ANT} & \multicolumn{2}{c|}{POLSKA}                         & \multicolumn{2}{c|}{PDH}    \\ \cline{2-9} 
Size                      & 100                      & 300                      & 100                    & 300  & 100                      & 300                      & 100                    & 300  \\ \hline
B\&L                  & 22.47        & 70.37       & 20.63       & 59.17       & 651.76      & 1965.00      & 298.26     & 896.00      \\
SPOT                  & 20.94        & 61.93       & 35.94       & 439.19      & 484.77      & 6281.35      & 223.53     & 2798.22     \\
SPO Forest            & 17.26        & 73.05       & 22.07       & 369.21      & 980.32      & 4277.86      & 488.36     & 2125.27 \\  \hline
\end{tabular}}
\caption{Average runtime (in $s$) of MCFP and MCVC on real-life data}
\label{table:MCFP_runtime}
\end{table}

The results show that the proposed methods have similar runtimes as SPOT and SPO Forest. The only exception is that the runtimes of SPOT and SPO Forest become quite large in G\'{E}ANT when the total set size is 300, while our proposed methods keep reasonable runtimes.

\section{Summary}
Given a $\texttt{ReSolve}$ function for a recursively or iteratively solvable problem, we propose a systematic approach to synthesize a $\texttt{ReLearn}$ function for learning with the regret loss, by replacing operators in $\texttt{ReSolve}$ with their piecewise linear counterparts.
Our proposal is methodical and straightforward to implement.
Furthermore, our framework encompasses a wide class of recursive algorithms (\verb~ReSolve~ functions), as demonstrated by our case studies.
Most importantly, B\&L empirically achieves the lowest regrets against classical machine learning and contemporary Predict+Optimize algorithms with runtime comparable with the latter.

\bibliographystyle{named}
\bibliography{ijcai22}

\begin{thebibliography}{}

\bibitem[\protect\citeauthoryear{Buitinck \bgroup \em et al.\egroup
  }{2013}]{sklearn_api}
Lars Buitinck, Gilles Louppe, Mathieu Blondel, Fabian Pedregosa, Andreas
  Mueller, Olivier Grisel, Vlad Niculae, Peter Prettenhofer, Alexandre
  Gramfort, Jaques Grobler, Robert Layton, Jake VanderPlas, Arnaud Joly, Brian
  Holt, and Ga{\"{e}}l Varoquaux.
\newblock {API} design for machine learning software: experiences from the
  scikit-learn project.
\newblock In {\em ECML PKDD Workshop: Languages for Data Mining and Machine
  Learning}, pages 108--122, 2013.

\bibitem[\protect\citeauthoryear{Cormen \bgroup \em et al.\egroup
  }{2009}]{cormen2009algorithm}
Thomas~H. Cormen, Charles~E. Leiserson, Ronald~L. Rivest, and Clifford Stein.
\newblock {\em Introduction to Algorithms, Third Edition}.
\newblock The MIT Press, 3rd edition, 2009.

\bibitem[\protect\citeauthoryear{Demirovi{\'c} \bgroup \em et al.\egroup
  }{2019a}]{demirovic2019investigation}
Emir Demirovi{\'c}, Peter~J Stuckey, James Bailey, Jeffrey Chan, Chris Leckie,
  Kotagiri Ramamohanarao, and Tias Guns.
\newblock An investigation into prediction+ optimisation for the knapsack
  problem.
\newblock In {\em International Conference on Integration of Constraint
  Programming, Artificial Intelligence, and Operations Research}, pages
  241--257. Springer, 2019.

\bibitem[\protect\citeauthoryear{Demirovi\'{c} \bgroup \em et al.\egroup
  }{2019b}]{demirovic2019predict+}
Emir Demirovi\'{c}, Peter~J Stuckey, James Bailey, Jeffrey Chan, Christopher
  Leckie, Kotagiri Ramamohanarao, and Tias Guns.
\newblock Predict+optimise with ranking objectives: Exhaustively learning
  linear functions.
\newblock {\em IJCAI-19}, pages 1078--1085, 2019.

\bibitem[\protect\citeauthoryear{Demirovi\'{c} \bgroup \em et al.\egroup
  }{2020}]{demirovic2020dynamic}
Emir Demirovi\'{c}, Peter~J Stuckey, Tias Guns, James Bailey, Christopher
  Leckie, Kotagiri Ramamohanarao, and Jeffrey Chan.
\newblock Dynamic programming for predict+optimise.
\newblock In {\em AAAI}, pages 1444--1451, 2020.

\bibitem[\protect\citeauthoryear{Elmachtoub and
  Grigas}{2020}]{elmachtoub2020smart}
Adam~N Elmachtoub and Paul Grigas.
\newblock Smart ``predict, then optimize".
\newblock {\em Management Science}, accepted, 2020.

\bibitem[\protect\citeauthoryear{Elmachtoub \bgroup \em et al.\egroup
  }{2020}]{elmachtoub2020decision}
Adam~N. Elmachtoub, Jason Cheuk~Nam Liang, and Ryan McNellis.
\newblock Decision trees for decision-making under the predict-then-optimize
  framework.
\newblock In {\em Proceedings of the 37th International Conference on Machine
  Learning, {ICML} 2020, 13-18 July 2020, Virtual Event}, volume 119 of {\em
  Proceedings of Machine Learning Research}, pages 2858--2867. {PMLR}, 2020.

\bibitem[\protect\citeauthoryear{Guler \bgroup \em et al.\egroup
  }{2022}]{ali2022divide}
Ali~Ugur Guler, Emir Demirovi\'{c}, Jeffrey Chan, James Bailey, Christopher
  Leckie, and Peter~J Stuckey.
\newblock A divide and conquer algorithm for predict+optimize with non-convex
  problems.
\newblock In {\em Proceedings of the AAAI Conference on Artificial
  Intelligence}, 2022.

\bibitem[\protect\citeauthoryear{LLC}{2018}]{WinNT}
MultiMedia LLC.
\newblock Geant topology map dec2018 copy.
\newblock
  \url{https://www.geant.org/Resources/Documents/GEANT_Topology_Map_December_2018.pdf},
  2018.
\newblock Accessed: 2020-09-10.

\bibitem[\protect\citeauthoryear{Lucerna \bgroup \em et al.\egroup
  }{2009}]{lucerna2009efficiency}
Diego Lucerna, Nicola Gatti, Guido Maier, and Achille Pattavina.
\newblock On the efficiency of a game theoretic approach to sparse regenerator
  placement in wdm networks.
\newblock In {\em GLOBECOM 2009-2009 IEEE Global Telecommunications
  Conference}, pages 1--6. IEEE, 2009.

\bibitem[\protect\citeauthoryear{Mandi and Guns}{2020}]{mandi2020interior}
Jayanta Mandi and Tias Guns.
\newblock Interior point solving for lp-based prediction+optimisation.
\newblock In {\em Advances in Neural Information Processing Systems}, page
  accepted, 2020.

\bibitem[\protect\citeauthoryear{Orlowski \bgroup \em et al.\egroup
  }{2007}]{SNDlib10}
S.~Orlowski, M.~Pi{\'o}ro, A.~Tomaszewski, and R.~Wess{\"a}ly.
\newblock {SNDlib} 1.0--{S}urvivable {N}etwork {D}esign {L}ibrary.
\newblock In {\em Proceedings of the 3rd International Network Optimization
  Conference (INOC 2007), Spa, Belgium}, April 2007.
\newblock {http://sndlib.zib.de, extended version accepted in Networks, 2009.}

\bibitem[\protect\citeauthoryear{Perron and Furnon}{2019}]{ortools}
Laurent Perron and Vincent Furnon.
\newblock Or-tools, 2019.

\bibitem[\protect\citeauthoryear{Von~Mohrenschildt}{1998}]{von1998normal}
Martin Von~Mohrenschildt.
\newblock A normal form for function rings of piecewise functions.
\newblock {\em Journal of Symbolic Computation}, 26(5):607--619, 1998.

\bibitem[\protect\citeauthoryear{Waissi}{1994}]{waissi1994network}
Gary~R Waissi.
\newblock Network flows: Theory, algorithms, and applications, 1994.

\bibitem[\protect\citeauthoryear{Wilder \bgroup \em et al.\egroup
  }{2019}]{wilder2019melding}
Bryan Wilder, Bistra Dilkina, and Milind Tambe.
\newblock Melding the data-decisions pipeline: Decision-focused learning for
  combinatorial optimization.
\newblock In {\em Proceedings of the AAAI Conference on Artificial
  Intelligence}, volume~33, pages 1658--1665, 2019.

\end{thebibliography}

\end{document}